\documentclass{article}
\usepackage{spconf,amsmath,graphicx,hyperref}
\usepackage{algorithm}
\usepackage{algpseudocode}  % 推荐替代 algorithmic
\usepackage{tabularx}
\usepackage{float}
\usepackage[section]{placeins} % 或者不带 [section]
\usepackage{subcaption}  % 导言区加上这个
\usepackage{cleveref}
\usepackage{amssymb}
\title{GRASP: GRoup-shApley feature Selection for Patients}
\ifdefined\HCode
  \author{
  Yuheng Luo$^{1}$,
  Shuyan Li$^{2}$,
  Zhong Cao$^{3,*}$\\
  $^{1}$Chinese Academy of Medical Sciences \& Peking Union Medical College, Beijing, China.\\
  $^{2}$Queen's University, Belfast, United Kingdom.
  $^{3}$Heidelberg University, Heidelberg, Germany.\\
  \texttt{*Email: zhong.cao@uni-heidelberg.de}
  }
\else
  \name{Yuheng Luo$^{1}$, Shuyan Li$^{2}$, Zhong Cao$^{3,*}$}
  \address{
  $^{1}$Chinese Academy of Medical Sciences \& Peking Union Medical College, Beijing, China\\
  $^{2}$Queen's University, Belfast, United Kingdom.
  $^{3}$Heidelberg University, Heidelberg, Germany.\\
  \smallskip
  \texttt{*Email: zhong.cao@uni-heidelberg.de}
  }
\fi

\begin{document}
%\ninept
%
\maketitle
\begin{abstract}

% Feature selection remains a major challenge in medical prediction, where existing approaches such as LASSO often lack robustness and interpretability. We introduce GRASP, a novel framework that couples Shapley value–driven attribution with group-$L_{21}$ regularization to extract compact and predictive feature sets. GRASP first distills group-level importance scores from a pre-trained tree model via SHAP, then enforces structured sparsity through group-$L_{21}$–regularized logistic regression, yielding stable and interpretable selections. Extensive comparisons with LASSO, SHAP, and deep learning–based methods show that GRASP consistently delivers comparable or superior predictive accuracy, while identifying fewer, less redundant, and more stable features.

Feature selection remains a major challenge in medical prediction, where existing approaches such as LASSO often lack robustness and interpretability. We introduce GRASP, a novel framework that couples Shapley value driven attribution with group $L_{21}$ regularization to extract compact and non-redundant feature sets. GRASP first distills group level importance scores from a pretrained tree model via SHAP, then enforces structured sparsity through group $L_{21}$ regularized logistic regression, yielding stable and interpretable selections. Extensive comparisons with LASSO, SHAP, and deep learning based methods show that GRASP consistently delivers comparable or superior predictive accuracy, while identifying fewer, less redundant, and more stable features.

\end{abstract}
\section{Introduction}
\label{sec:intro}

With the growth of electronic health records, medical imaging, and wearable devices, healthcare systems are generating vast amounts of phenotypic data that capture patients’ clinical characteristics, disease manifestations, and treatment responses. These data offer great potential for precision medicine, but their high dimensionality and noise pose major challenges for knowledge discovery \cite{wang2019deep}. Feature selection is essential as it reduces computational costs by enabling the extraction of the most informative variables.
% \cite{topol2014individualized} \cite{guyon2003introduction}

Feature selection methods are typically categorized as filter, wrapper, and embedded approaches \cite{guyon2008feature,gui2016feature,li2017feature}. Filter methods score features using statistical or information-theoretic criteria, offering speed and scalability but often ignoring inter-feature interactions and redundancy \cite{chandrashekar2014survey}. Wrappers evaluate feature subsets via model performance, achieving higher accuracy at significant computational cost \cite{kohavi1997wrappers,tang2014feature}. Embedded methods integrate selection into model training, with L1-norm regularization (e.g., LASSO \cite{tibshirani1996regression}, Graphical Lasso \cite{rey2025enhancing}) and tree-based models \cite{chen2016xgboost}) being widely applied. Recent studies have also focused on developing specific scoring functions to evaluate the consistency and reliability of feature rankings \cite{marinescuscoring}. Furthermore, another study \cite{san2022exhaustive} that exhaustively compares selection methods reveals that the choice of a selector can dramatically affect the final set of variables. Despite advances, existing approaches remain unstable, redundant, and often lack biological interpretability \cite{khaire2022stability}. Moreover, most studies emphasize predictive accuracy while overlooking interpretability and clinical relevance—two requirements critical for medical adoption \cite{ghassemi2020review}. Post-hoc explanation tools such as SHAP provide insights, but their outputs may difficult to align with disease-specific mechanisms.
% ,ke2017lightgbm
% \begin{figure}[htb]
%   \centering
%   \centerline{\includegraphics[width=8.5cm]{fig0.pdf}}
% %  \vspace{2.0cm}
%   % \centerline{(a) Result 1}\medskip%
% \caption{Main contribution of this paper.}
% \label{fig:contribution}
% %
% \end{figure}

To address these challenges, we propose GRASP (GRoup-shApley feature Selection for Patients), an interpretable feature selection algorithm that integrates cross-entropy with SHAP-based attribution into a penalized framework. Unlike prior methods that separate prediction and interpretability, GRASP unifies both to achieve stable selection and clinically meaningful insights. Our main contributions are:

 \vspace{-0.2 cm}
\begin{enumerate}
\item \textbf{Interpretable framework.} We are the first to couple SHAP-based feature attribution with regularization, enabling optimization algorithms to preserve clinically interpretable features.
 \vspace{-0.2 cm}
\item \textbf{Grouped selection mechanism.} We design a group-based strategy that simplifies models and enhances stability and generalization.
 \vspace{-0.2 cm}
\item \textbf{Comprehensive evaluation.} This study evaluates feature-selection methods beyond accuracy, offering valuable guidance for the future development of feature selection techniques.
\end{enumerate}

\section{Method}
\label{sec:method}

\subsection{Overview}
\begin{figure}[htb]
  \centering
  \centerline{\includegraphics[width=8.5cm]{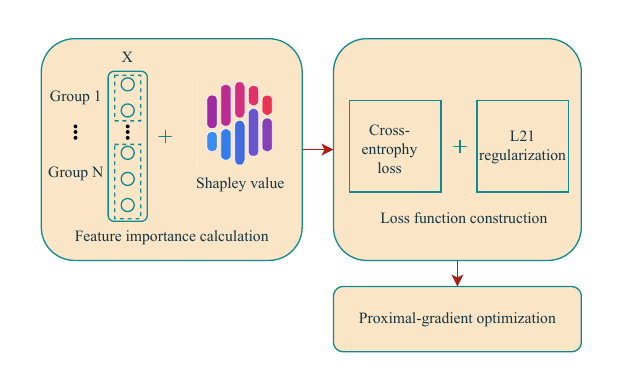}}
 \vspace{-0.5 cm}
  % \centerline{(a) Result 1}\medskip%
\caption{Workflow of GRASP algorithm.The procedure begins with feature importance calculation by combining input feature groups with Shapley values. These importance scores are then incorporated into a loss function that integrates $L_{21}$ loss and group-wise $L_{21}$ regularization. The final optimization is performed using a proximal-gradient algorithm.}
\label{fig:workflow}
\end{figure}

We propose GRASP, a feature selection method that integrates model-derived attributions with group-$L_{21}$ regularized logistic regression, optimized via a proximal-gradient algorithm with Armijo backtracking \cite{armijo1966minimization}. The procedure consists of: (1) feature importance calculation; (2) loss function construction; and (3) proximal-gradient optimization. Formally, let $X \in  \mathbb{R}^{n\times p}$ be the design matrix with binary outcome $y\in{0,1}^n$, and partition features into $G$ disjoint groups $\mathcal{G}={g_1,\dots,g_G}$. $\beta_g$ denotes the subvector for group $g$ with Euclidean norm $||\beta_g||_2$. Feature importance is $\phi_j$, and aggregated group importance is $s_g$. Fig~\ref{fig:workflow} shows the main workflow of GRASP.

% 统一说明NHANES里面用validation，UKB作为test数据集
\subsection{Feature importance calculation}
Given an XGBoost classifier trained on the training fold, compute SHAP values on the held-out validation fold $X^{\text{val}}$. For each feature $j$ and sample $i$ let $\mathrm{SHAP}_{ij}$ be the corresponding Shapley value, which is defined as:

\begin{equation}
\phi_j = \frac{1}{n_{\text{val}}}\sum_{i=1}^{n_{\text{val}}} \bigl|\text{SHAP}_{ij}\bigr|.
\end{equation}
For each group $g$ we aggregate by the mean of its member feature importances:
\begin{equation}
s_g = \frac{1}{|g|}\sum_{j\in g}\phi_j.
\end{equation}
The group feature selection strategy in this equation allows highly correlated features to enter or leave together to some extent, enhancing the stability of our feature selection method. Meanwhile, the incorporation of SHAP values ensures that the selected features are more contributive and, consequently, easier to interpret.

To obtain positive group penalty coefficients that decrease with greater importance, we transform $s_g$ into weights:
\begin{equation}
\tilde{\omega}_g = \exp\bigl(-s_g / \tau_0\bigr) + \varepsilon,\qquad
\omega_g = \frac{\tilde{\omega}_g}{\sum_{h=1}^{G}\tilde{\omega}_h},
\label{eq:omega_norm_revised}
\end{equation}
where $\tau_0$ is an optional scaling parameter (set to 1 by default) and $\varepsilon>0$ a small constant for numerical stability. This ensures $\omega_g>0$ and $\sum_g\omega_g=1$. Intuitively, groups with larger SHAP scores $s_g$ receive smaller $\omega_g$ and thus a reduced multiplicative penalty in the group regularizer.

\subsection{Loss function construction}
We then fit our loss function with a group $L_{21}$ penalty. The loss is defined as
% \begin{align}
% \mathcal{L}(\beta) = -\frac{1}{n}\sum_{i=1}^n\Bigl[y_i\log\sigma(x_i^\top\beta) \\
% & + (1-y_i)\log\bigl(1-\sigma(x_i^\top\beta)\bigr)\Bigr],
% \label{eq:logloss}
% \end{align}
\begin{equation}
\begin{aligned}
\mathcal{L}(\beta)
= -\frac{1}{n}\sum_{i=1}^n \Bigl[
&y_i\log\sigma(x_i^\top\beta) \\
&+ (1-y_i)\log\bigl(1-\sigma(x_i^\top\beta)\bigr)
\Bigr],
\end{aligned}
\label{eq:logloss}
\end{equation}

where $\sigma(z)=(1+e^{-z})^{-1}$ is the sigmoid. The regularized objective is

\begin{equation}
J(\beta) = \mathcal{L}(\beta) + \lambda \sum_{g=1}^G \omega_g \|\beta_g\|_2,
\label{eq:objective}
\end{equation}
with $\lambda>0$ the global regularization parameter and $\omega_g$ defined in Eq.~\eqref{eq:omega_norm_revised}. The penalty term $\Omega=\sum_g \omega_g \|\beta_g\|_2$ enforces group-level sparsity while allowing dense coefficients within selected groups. The key tuning parameter is the regularization weight $\lambda$. We adopt a two-step strategy inspired by Gordon’s Theorem \cite{vershynin2010introduction}: after normalization, we compute the empirical noise matrix $E = X^{\text{train}} - \mathbf{1}\bar X^\top$ and set $\lambda$ to $\mathrm{sd}(E)$.
% ,zhao2006model
\subsection{Proximal-gradient optimization}
We optimize Eq.~\eqref{eq:objective} using a proximal-gradient scheme. Given iterate $\beta^{(k)}$ and step-size $t_k>0$, the update is
\begin{align}
v_{k+1} &= \beta^{(k)} - t_k \nabla\mathcal{L}\bigl(\beta^{(k)}\bigr), \\
\beta^{(k+1)} &= \mathrm{prox}_{t_{k}\lambda\Omega}(v_{k+1}),
\end{align}
where the proximal operator for the group $L_{21}$ penalty acts group-wise:
\begin{equation}
\mathrm{prox}_{\tau\omega_g\|\cdot\|_2}(v_g) =
\begin{cases}
\displaystyle \left(1 - \frac{\tau\omega_g}{\|v_g\|_2}\right)v_g, & \text{if } \|v_g\|_2 > \tau\omega_g,\\[6pt]
\mathbf{0}, & \text{otherwise},
\end{cases}
\label{eq:prox}
\end{equation}
with $\tau = t_k\lambda$. 

An Armijo-type backtracking line-search is employed to select the step-size $t_k$. Specifically, based on the descent lemma for smooth functions, we accept $t_k$ if
\begin{align}
\mathcal{L}(\beta^{(k+1)}) \le 
& \ \mathcal{L}(\beta^{(k)}) 
+ \nabla\mathcal{L}(\beta^{(k)})^\top(\beta^{(k+1)}-\beta^{(k)}) \nonumber \\
& + \frac{1}{2t_k}\|\beta^{(k+1)}-\beta^{(k)}\|_2^2,
\label{eq:armijo}
\end{align}
which ensures sufficient decrease of the objective.
If the condition is not satisfied, the step-size is reduced by $t_k \leftarrow \alpha t_k$ with $\alpha\in(0,1)$ (we use $\alpha=0.5$), and the test is repeated. The algorithm terminates when $||\beta^{(k+1)}-\beta^{(k)}||_2 < \varepsilon$ or a maximum number of iterations is reached.
% 伪代码表格
% \begin{algorithm}[t]
% \caption{Proximal-gradient optimization procedure}
% \label{alg:pg_groupl21}
% \begin{algorithmic}[1]
% \Require $X,y,\mathcal{G},\{\omega_g\}_{g=1}^G,\lambda,t_{\text{init}},\alpha,\varepsilon,K_{\max}$
% \State initialize $\beta^{(0)}\gets \mathbf{0}$, $t\gets t_{\text{init}}$
% \For{$k=0,1,\dots,K_{\max}-1$}
%   \State compute $g^{(k)}\gets\nabla\mathcal{L}(\beta^{(k)})$
%   \While{true}
%     \State $v \gets \beta^{(k)} - t\, g^{(k)}$
%     \State $\beta^{\text{cand}} \gets \mathrm{prox}_{t\lambda\Omega}(v)$
%     \State evaluate $\mathcal{L}(\beta^{\text{cand}})$
%     \If{Eq.~\eqref{eq:armijo} holds}
%       \State \textbf{break}
%     \Else
%       \State $t \gets \alpha t$
%     \EndIf
%   \EndWhile
%   \If{$\|\beta^{\text{cand}}-\beta^{(k)}\|_2 < \varepsilon$}
%     \State \Return $\beta^{\text{cand}}$
%   \EndIf
%   \State $\beta^{(k+1)} \gets \beta^{\text{cand}}$
% \EndFor
% \State \Return $\beta^{(k+1)}$
% \end{algorithmic}
% \end{algorithm}

Finally, we compute the group norms $||\beta_g||_2$ and retain the groups satisfying
\begin{equation}
\widehat{\mathcal{G}} = {g:\ ||\beta_g||_2 > 0}.
\end{equation}
All features outside the selected groups are discarded.

\section{Experiments}
\label{sec:experiments}

\begin{figure}[htbp]
% \begin{figure}[H]
  \centering
  \centerline{\includegraphics[width=8.5cm]{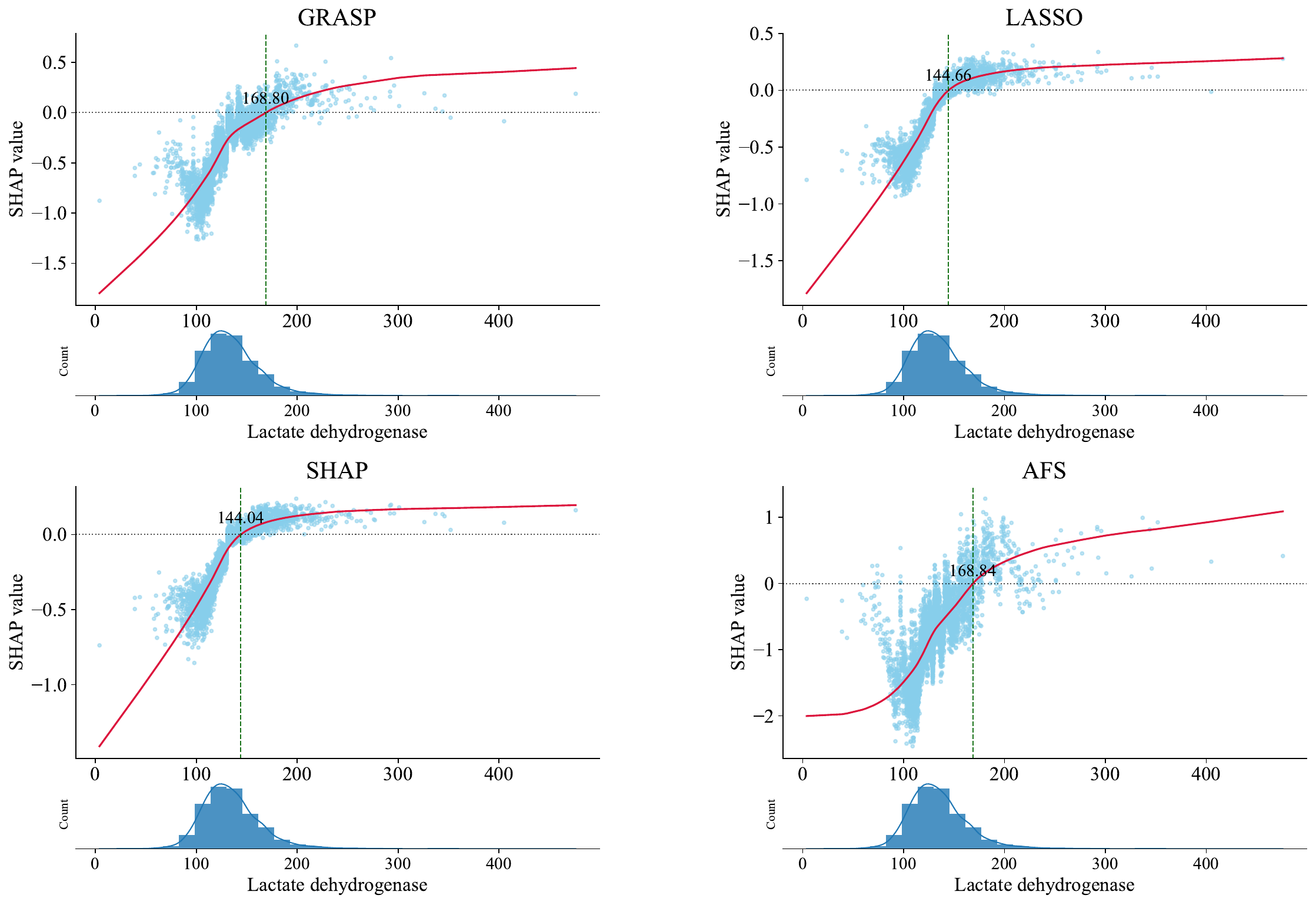}}
 % \vspace{-0.5cm}
  % \centerline{(a) Result 1}\medskip%
\caption{Main effect plots of Lactate dehydrogenase (LDH) using overlapping feature sets from GRASP, LASSO, SHAP and AFS. Red curves depict LOWESS curves and blue dots
show Shapley values. Histograms indicate LDH distribution, and dashed lines mark estimated thresholds across methods.}
\label{fig:fig_shap}
% \vspace{-0.3cm}
%
\end{figure}

% width=8.5cm
% height=6cm
% 没地方了就只能把这张图删掉算了
\begin{figure}[htbp]
% \begin{figure}[H]
  \centering
  \centerline{\includegraphics[width=8cm]{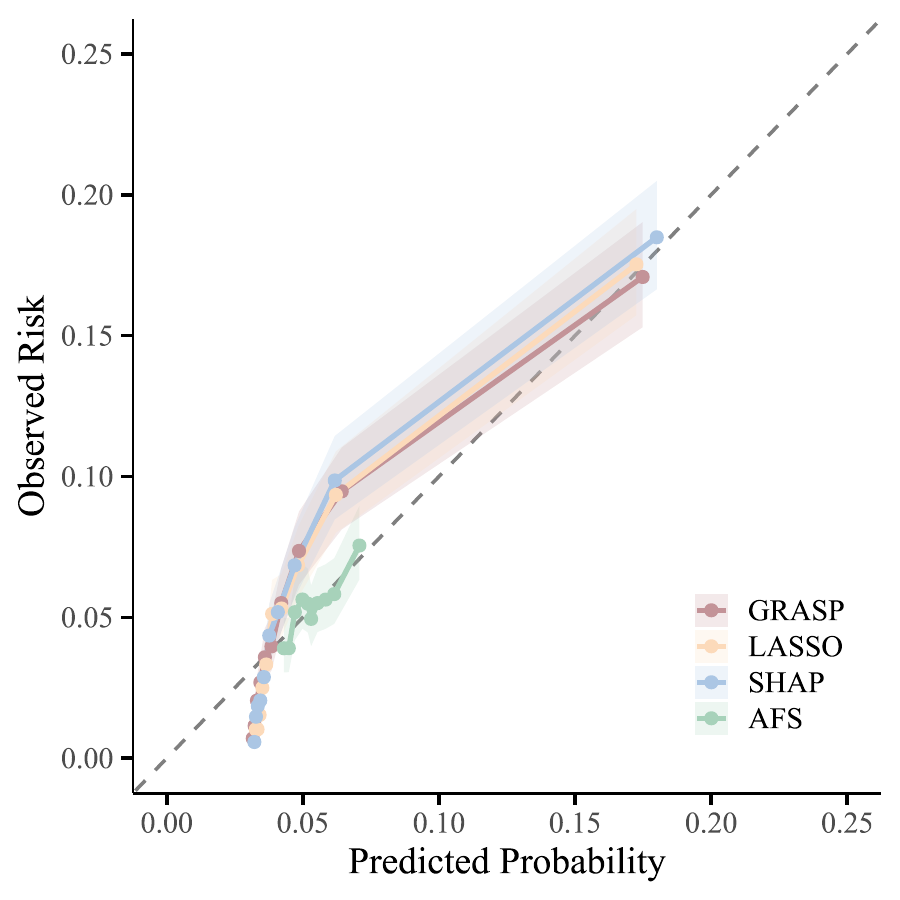}}
 % \vspace{-0.5cm}
  % \centerline{(a) Result 1}\medskip%
\caption{Calibration curve comparing predicted probabilities and observed risks for models based on feature sets from GRASP, LASSO, SHAP, and AFS. All results are calibrated using the Platt Scaling method. The dashed line indicates perfect calibration, with shaded areas representing 95\% confidence intervals.}
\label{fig:fig_calib}
%
% \vspace{-0.3cm}
\end{figure}

\subsection{Experiments Setup}
% 无括号版结果

\textbf{Datasets.} 
We used two datasets. The development dataset was NHANES (National Health and Nutrition Examination Survey) \cite{nguyen2023harmonized}, a biennial US survey combining interviews, physical examinations, and laboratory tests. We used 1999–2014 cycles, restricted to adults ($\geq$20 years) with mortality follow-up through December 31, 2019. Variables with $>$10\% missingness were removed. All-cause mortality were obtained from the linked mortality file. UK Biobank (UKB) \cite{sudlow2015uk} served as external validation (accessed under Application Number 240523), with 15,082 participants and confirmed 5-year mortality (before July 2024). To assess generalizability of features selected by different feature selection methods, 76 variables (86 after encoding) overlapped between NHANES and UKB were used. For more details regarding data processing and targeted patients, please visit \url{https://github.com/yhluo7/GRASP}.

\textbf{Baselines.} 
We compared GRASP with LASSO, SHAP method, and the AFS algorithm \cite{gui2019afs}. For LASSO, the penalty parameter was tuned by five-fold cross-validation. For SHAP, per-feature Shapley values were aggregated and features above the median value were retained. AFS was run with default settings. To avoid data leakage, preprocessing procedure was fitted on training fold and applied on validation fold: (1) one-hot encoding of categorical variables; (2) median and mode imputation; (3) z-score standardization of continuous features.

\textbf{Evaluation metrics.} We assessed feature selection by predictability (Accuracy, F1 score, and Matthews correlation coefficient (MCC)), redundancy (average Variance Inflation Factor (VIF), Shapley value \cite{lundberg2017unified} and Pearson correlation coefficient), and stability (Jaccard Index (JI) and the Adjusted Stability Measure (ASM) \cite{lustgarten2009measuring}). We applied 1,000 bootstrap iterations on the NHANES validation set and UKB dataset to obtain reliable estimates. We also use calibration curves and Kaplan–Meier survival curves to evaluate the risk estimation and stratification ability of different selected features by training a COX model on the UKB dataset.
% \cite{chicco2020advantages}
% \cite{chelvan2016study}

% \begin{figure}[htb]
% \centering
% \begin{subfigure}[b]{0.50\linewidth} % 改为两行两列排布
%   \centering
%   % 第一行
%   \begin{minipage}[b]{0.48\linewidth}
%     \centering
%     \includegraphics[width=\linewidth]{confer_UKB_KM_XGB_GRASP_curve}
%     \subcaption{GRASP}
%   \end{minipage}
%   \begin{minipage}[b]{0.48\linewidth}
%     \centering
%     \includegraphics[width=\linewidth]{confer_UKB_KM_XGB_LASSO_curve}
%     \subcaption{LASSO}
%   \end{minipage}
%   % 换行
%   \vskip\baselineskip
%   % 第二行
%   \begin{minipage}[b]{0.48\linewidth}
%     \centering
%     \includegraphics[width=\linewidth]{confer_UKB_KM_XGB_SHAP_curve}
%     \subcaption{SHAP}
%   \end{minipage}
%   \begin{minipage}[b]{0.48\linewidth}
%     \centering
%     \includegraphics[width=\linewidth]{confer_UKB_KM_XGB_AFS_curve}
%     \subcaption{AFS}
%   \end{minipage}
% \end{subfigure}
% \caption{Comparison of different feature selection methods using Kaplan-Meier curve. 
%   Confidence bands represent the 95\% confidence interval. 
%   The global log-rank test and test statistic are reported in each Kaplan–Meier plot.}
% \label{fig:fig2}
% \end{figure}

% 原来的table1统计图表
% \begin{table}[]
% \centering
% \caption{Statistics of the datasets.}  % 添加标题
% \label{tab:tab1}
% \begin{tabular}{llll}
% \hline
% Dataset & \# Survivor & \# Nonsurvivor & Feature in total \\ \hline
% NHANES & 4388 & 609 & 4740 \\
% UKB & 14246 & 836 & 9805 \\ \hline
% \end{tabular}
% \end{table}

\begin{table*}[t]
\centering
\caption{Comparison of feature predictive performance on NHANES and UKB datasets. Other results are statistically significant at the 5\% level compared with the GRASP method. The best score for each metric within a dataset is highlighted in bold. A dash indicates that the model predicted all samples as negative under the default probability threshold of 0.5.}
\label{tab:tab3}
\begin{tabularx}{\textwidth}{l l *{4}{X} *{4}{X}}
\hline
 & & \multicolumn{4}{c}{NHANES} & \multicolumn{4}{c}{UKB} \\
Model & Metric & GRASP & LASSO & SHAP & AFS & GRASP & LASSO & SHAP & AFS \\ \hline
% LR  & AUC      & 0.875 & 0.879 & \textbf{0.883} & 0.681 & 0.742 & 0.747 & \textbf{0.753} & 0.561 \\
LR  & Accuracy & 0.783 & \textbf{0.786} & 0.785 & 0.650 & \textbf{0.755} & 0.752 & 0.735 & 0.704 \\
LR  & F1       & 0.483 & \textbf{0.491} & 0.487 & 0.283 & 0.197 & 0.203 & \textbf{0.206} & 0.120 \\
LR  & MCC      & 0.436 & \textbf{0.448} & 0.440 & 0.155 & 0.170 & 0.181 & \textbf{0.191} & 0.050 \\
% RF  & AUC      & 0.875 & 0.875 & \textbf{0.881} & 0.685 & 0.745 & 0.740 & \textbf{0.756} & 0.559 \\
RF  & Accuracy & \textbf{0.890} & 0.884 & 0.883 & 0.878 & 0.946 & 0.946 & 0.945 & \textbf{0.946} \\
RF  & F1       & \textbf{0.226} & 0.120 & 0.093 & ——   & \textbf{0.016} & 0.007 & 0.009 & ——   \\
RF  & MCC      & \textbf{0.297} & 0.206 & 0.186 & ——   & 0.045 & 0.032 & \textbf{0.048} & ——   \\
% XGB & AUC      & 0.881 & 0.880 & \textbf{0.887} & 0.684 & 0.743 & 0.745 & \textbf{0.757} & 0.547 \\
XGB & Accuracy & \textbf{0.897} & 0.895 & 0.886 & 0.878 & 0.942 & 0.940 & 0.937 & \textbf{0.946} \\
XGB & F1       & \textbf{0.437} & 0.431 & 0.393 & ——   & 0.143 & 0.137 & \textbf{0.152} & ——   \\
XGB & MCC      & \textbf{0.416} & 0.406 & 0.356 & ——   & 0.142 & \textbf{0.150} & 0.142 & ——   \\ \hline
\end{tabularx}
\end{table*}

\begin{table}[!htbp]
\centering
\caption{Comparison of different feature selection methods and redundancy on the NHANES dataset. For LASSO and SHAP, the selected features exhibited multicollinearity, therefore the corresponding results are denoted with a dash. P\textless
0.05:*,P\textless0.01:**,P\textless0.001:***.}  % 添加标题
\label{tab:tab2}
\begin{tabularx}{\columnwidth}{lXXXX}
\hline
Metric & GRASP & LASSO & SHAP & AFS \\ \hline
Feature number & \textbf{23} & 44** & 43*** & 59 \\
% Jaccard index & \textbf{0.736} & 0.706 & 0.815** & 0.527 \\
Adjusted stability \\ measure & \textbf{0.593} & 0.382*** & 0.398*** & 0.258*** \\
Shapley Value & \textbf{0.101} & 0.074*** & 0.085** & 0.054*** \\
VIF & \textbf{2.942} & —— & —— & 8.201 \\
Correlation & 0.070 & 0.079** & 0.082** & \textbf{0.064} \\ \hline
\end{tabularx}
\vspace{-0.65cm}
\end{table}

\textbf{Implementation details.} We first split the NHANES dataset into training and validation sets with an 80:20 ratio. For each feature selection method, we restricted the analysis to the 76 features shared with UKB and performed five-fold partitioning within the NHANES training set. Features consistently selected across all five folds were retained as the final feature set for that method. Based on the selected features, we trained logistic regression, random forests and XGBoost models, tuning their hyperparameters on the training set using Optuna \cite{akiba2019optuna}. Model performance was then evaluated on the validation set and UKB through 1,000 bootstrap resamples to compare the ability of feature sets.

\subsection{Performance evaluation}

\begin{figure}[htbp]
% \vspace{-1.5cm}
\centering
\begin{subfigure}[b]{0.48\linewidth}
    \centering
    \includegraphics[width=\linewidth]{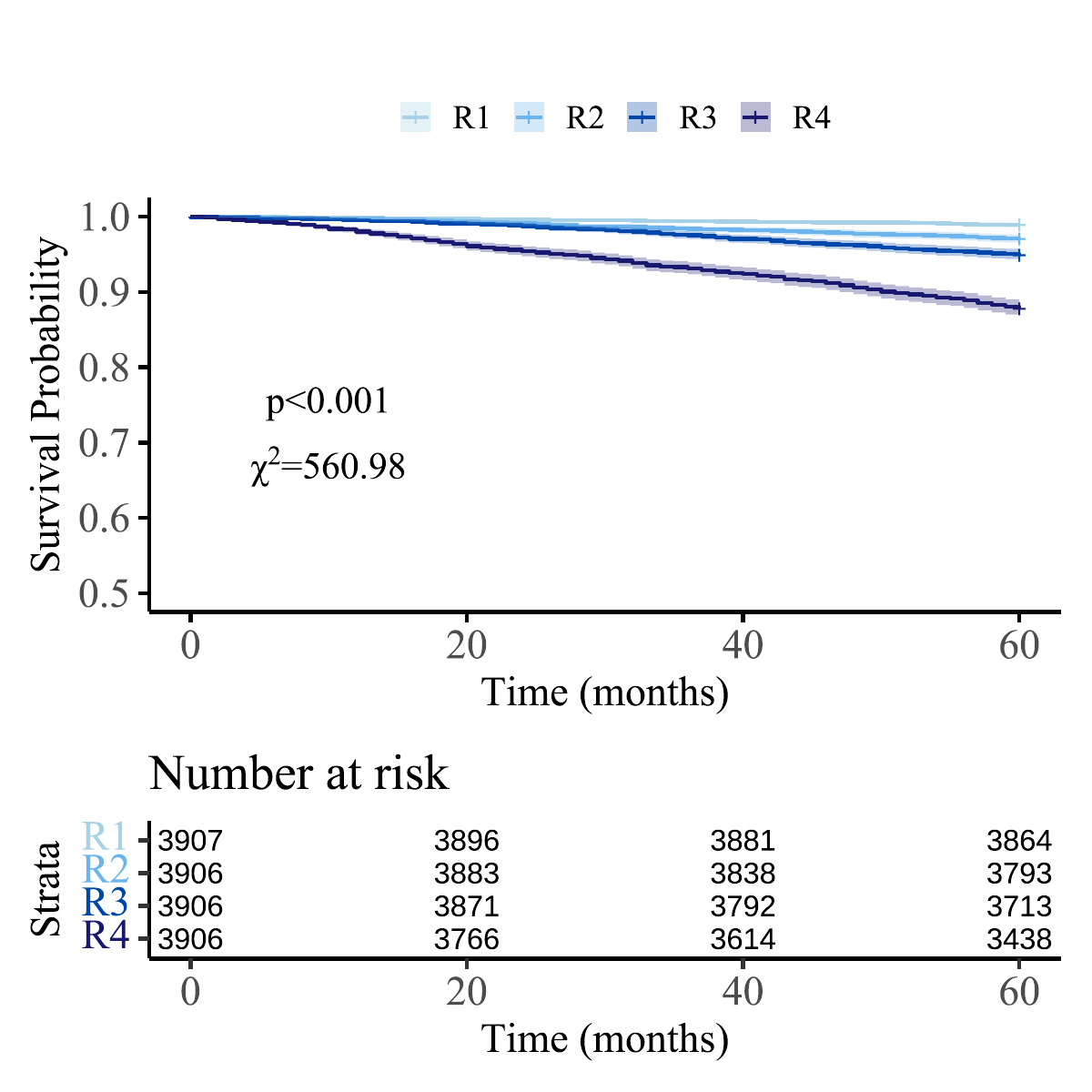}
    \subcaption{GRASP}
\end{subfigure}
\begin{subfigure}[b]{0.48\linewidth}
    \centering
    \includegraphics[width=\linewidth]{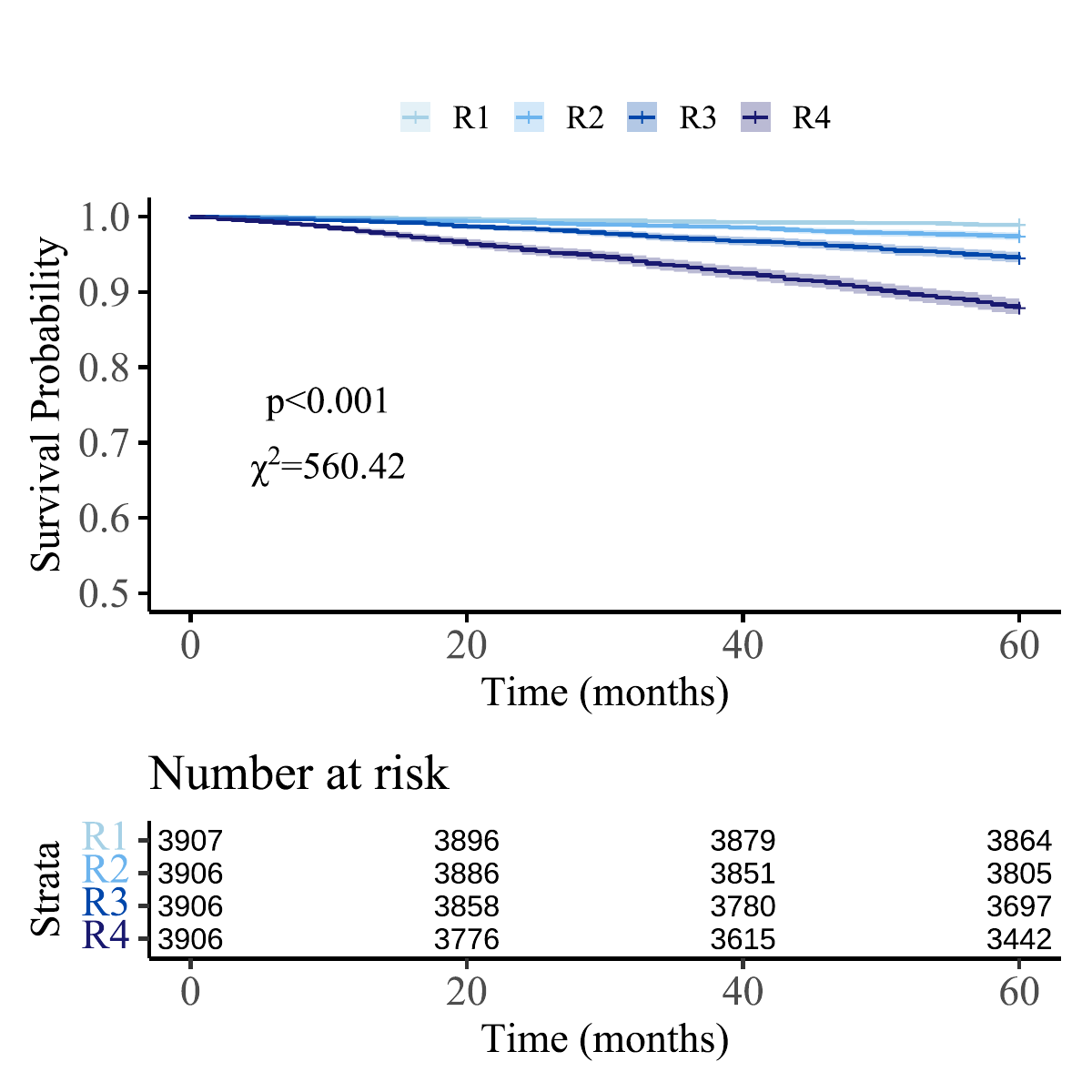}
    \subcaption{LASSO}
\end{subfigure}

\begin{subfigure}[b]{0.48\linewidth}
    \centering
    \includegraphics[width=\linewidth]{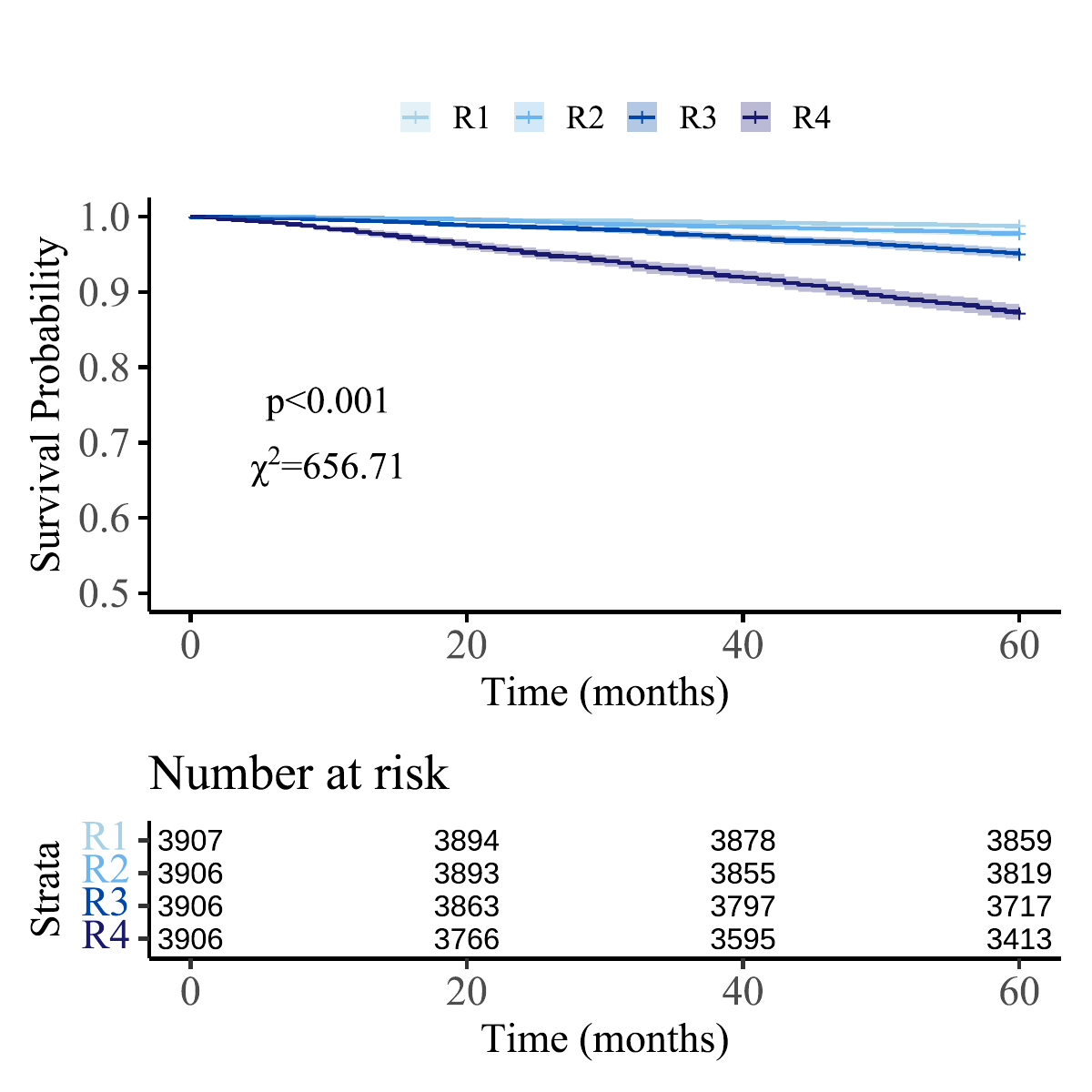}
    \subcaption{SHAP}
\end{subfigure}
\begin{subfigure}[b]{0.48\linewidth}
    \centering
    \includegraphics[width=\linewidth]{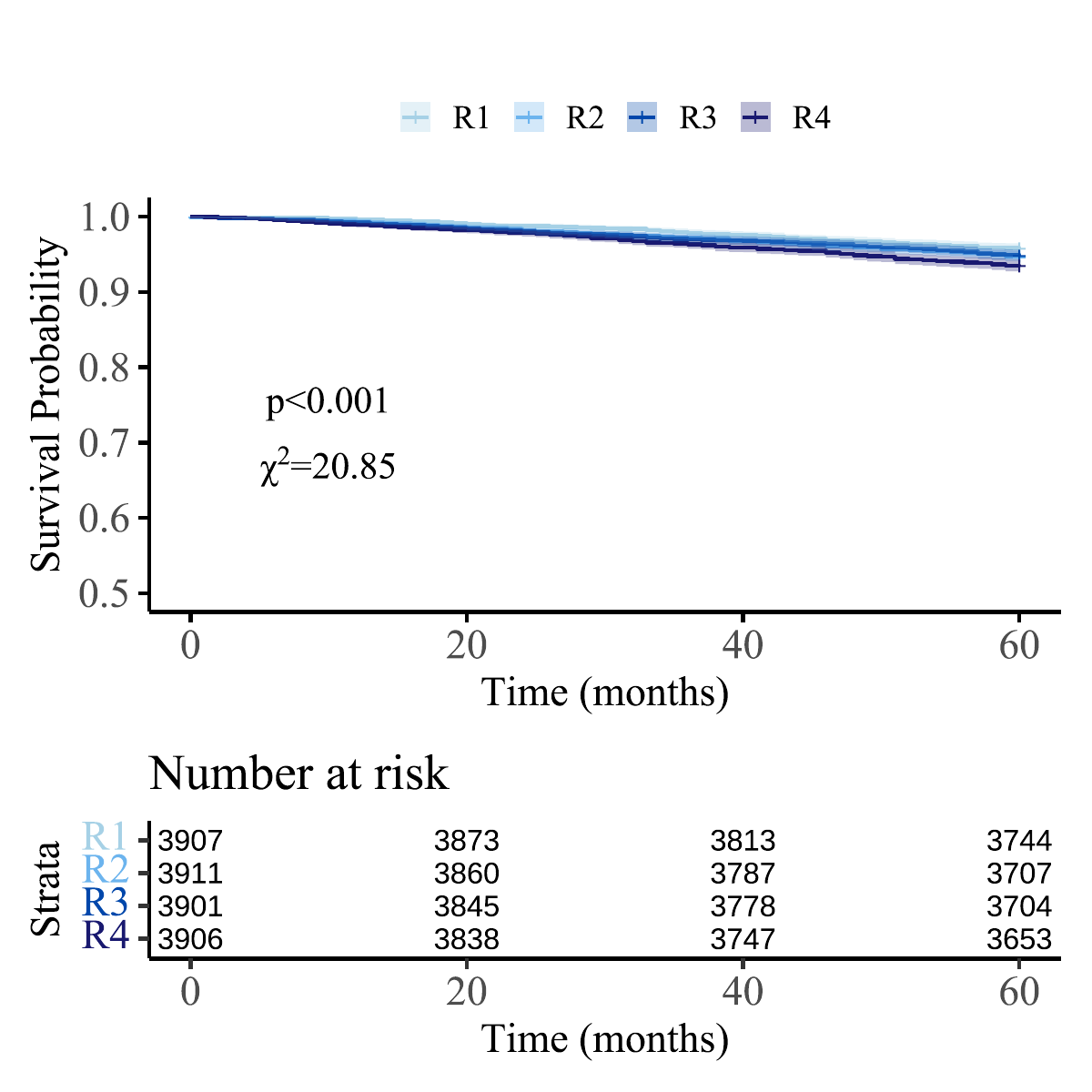}
    \subcaption{AFS}
\end{subfigure}

\caption{Comparison of different feature selection methods using Kaplan-Meier curves in UKB dataset. 
Confidence bands represent the 95\% confidence interval. 
The global log-rank test and test statistic are reported in each Kaplan–Meier plot.}
\label{fig:fig_km}
% 此处用来调整图片的垂直间隔！
% \vspace{-1cm}
\end{figure}

Results in \Cref{tab:tab2,tab:tab3} and \Cref{fig:fig_shap,fig:fig_km,fig:fig_calib} highlight that GRASP  identified the least feature set (23 features on average) with the highest stability (0.593) and lowest redundancy (VIF = 2.942), effectively avoiding the multicollinearity observed in LASSO and SHAP. Moreover, features selected by GRASP maintained predictive performance comparable to larger feature sets across NHANES and UKB. Calibration and  Kaplan-Meier analyses further confirmed its robustness. GRASP’s calibration curve aligned more closely with observed risks in high-risk groups, and its Kaplan-Meier stratification achieved equal or better discrimination than two other methods. For interpretability, focusing on  Lactate dehydrogenase (LDH, selected by all methods), GRASP produced a LOWESS curve whose transition point (168.8 U/L) was closest to the clinical cutoff reported by Guo et al. (315 U/L)\cite{guo2025association}, compared with 144.66 U/L for LASSO, 144.04 U/L for SHAP. AFS shares a similar threshold (168.84 U/L), but shows a sharper SHAP increase before it. Overall, GRASP aligns best with prior clinical evidence and yields the most interpretable features.

\section{Conclusion}
\label{sec:conclusion}
We develop a feature-selection method that combines $L_{21}$ norm with SHAP-based interpretability. Experiments on real-world datasets confirm its competitive performance compared with existing feature selection methods. Future studies could improve efficiency on high-dimensional datasets.
% \begin{figure}[H]

\vfill\pagebreak

% \section{funding acknowledgments}
\section{Acknowledgment}
\label{sec:ACKNOWLEDGMENT}
The work was supported by the Noncommunicable Chronic Diseases–National Science and Technology Major Project (Project Number 2023ZD0506000). The funders had no role in study design, data collection and analysis, decision to publish, or preparation of the manuscript.

\section{Compliance with ethical standards}
\label{sec:Compliance with ethical standards}
% This retrospective study used de-identified human participant data from NHANES (publicly available) and the UK Biobank (accessed under application number 240523). The original studies obtained ethics approval and informed consent. No additional ethical approval was required for this secondary analysis.
This study was conducted retrospectively using human subject data made available publicly by NHANES and the UK Biobank (accessed under
application number 240523). Ethical approval was not required as confirmed by the license attached with the public data.

%\vfill
%\pagebreak

% 这里才是换栏的核心部分！不需要换栏直接注释掉
% \vfill\pagebreak

% \section{REFERENCES}
% \label{sec:refs}

% List and number all bibliographical references at the end of the
% paper. The references can be numbered in alphabetic order or in
% order of appearance in the document. When referring to them in
% the text, type the corresponding reference number in square
% brackets as shown at the end of this sentence \cite{C2}. An
% additional final page (the fifth page, in most cases) is
% allowed, but must contain only references to the prior
% literature.

% Please follow the IEEE Citation Guidelines, \url{https://ieee-dataport.org/sites/default/files/analysis/27/IEEE\%20Citation\%20Guidelines.pdf} for formatting of references.

% 注意确认清楚能不能用small?实在不行就继续删除参考文献
\bibliographystyle{IEEEbib}
\bibliography{strings,refs}

% \begin{small}
% \bibliographystyle{IEEEbib}
% \bibliography{strings,refs}
% \end{small}

\end{document}